\documentclass[10pt,twocolumn,letterpaper]{article}

\usepackage[final]{cvpr}      

\usepackage{multirow}
\usepackage{adjustbox}
\usepackage{tabularx} 
\usepackage{siunitx}

\usepackage{algorithm}
\usepackage{algpseudocode}
\usepackage{amsmath}
\usepackage{amssymb}

\definecolor{cvprblue}{rgb}{0.21,0.49,0.74}
\usepackage[pagebackref,breaklinks,colorlinks,allcolors=cvprblue]{hyperref}

\def\paperID{***} 
\def\confName{CVPR}
\def\confYear{2026}

\title{CoopDiff: A Diffusion-Guided Approach for Cooperation under Corruptions}

\author{
Gong Chen$^{1}$ \quad
Chaokun Zhang$^{2}$\thanks{Corresponding author.} \quad
Pengcheng Lv$^{3}$ 
\\
$^{1}$School of Computer Science and Technology, Tianjin University  \quad \\
$^{2}$School of Cybersecurity, Tianjin University  \quad
$^{3}$School of Future Technology, Tianjin University 
}

\begin{document}

\maketitle

\begin{abstract}


Cooperative perception lets agents share information to expand coverage and improve scene understanding. However, in real-world scenarios, diverse and unpredictable corruptions undermine its robustness and generalization. To address these challenges, we introduce CoopDiff, a diffusion-based cooperative perception framework that mitigates corruptions via a denoising mechanism. CoopDiff adopts a teacher-student paradigm: the Quality-Aware Teacher performs voxel-level early fusion with Quality of Interest weighting and semantic guidance, then produces clean supervision features via a diffusion denoiser. The Dual-Branch Diffusion Student first separates ego and cooperative streams in encoding to reconstruct the teacher's clean targets. And then, an Ego-Guided Cross-Attention mechanism facilitates balanced decoding under degradation by adaptively integrating ego and cooperative features. We evaluate CoopDiff on two constructed multi-degradation benchmarks, OPV2Vn and DAIR-V2Xn, each incorporating six corruption types, including environmental and sensor-level distortions. Benefiting from the inherent denoising properties of diffusion, CoopDiff consistently outperforms prior methods across all degradation types and lowers the relative corruption error. Furthermore, it offers a tunable balance between precision and inference efficiency.

\end{abstract}

\begin{figure}[tp] 
  \centering 
  \includegraphics[width=0.95\linewidth]{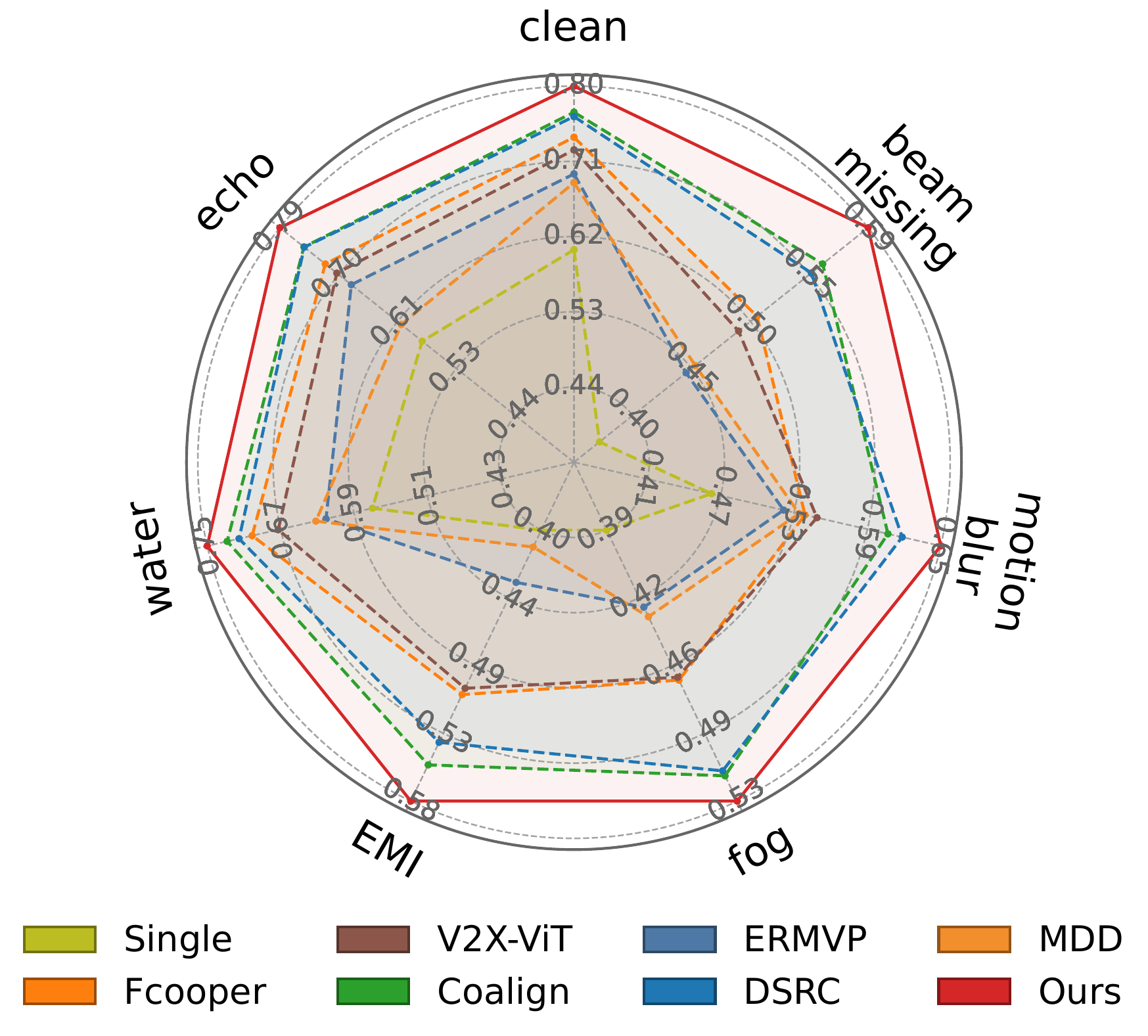}
  \caption{Overall performance comparison of average results on eight state-of-art methods across six corruption types.}
  \label{fig:abstract} 
\end{figure}

\section{Introduction}

Cooperative perception enables multiple agents to enhance their environmental understanding through information sharing \cite{li2024bevformer, yang2023bevformer}, thereby extending the perceptual field and improving accuracy. Owing to its efficient data fusion and complementary perspectives, it has emerged as a key technology for improving the performance and safety of autonomous driving systems \cite{survey1, survey2}. However, while cooperative perception can greatly strengthen system robustness under ideal conditions, real-world deployment remains challenging due to various sources of environmental noise \cite{RCPBench, xu2022v2xvit} and communication interference \cite{tang2025rocooper, v2xdgw, gong}.




These degradations significantly undermine cooperative perception. Generally, they can be broadly classified into two main categories.
The first is environmental data quality degradation \cite{tang2025rocooper, zhang2025dsrc}, such as from fog or echo reflections. This issue broadly reduces the signal-to-noise ratio of each agent's data. When this low-quality, dirty data is shared and fused across the network, noise can accumulate and potentially be amplified, thereby contaminating the final perception result. The second category is communication failure \cite{mrcnet, yang2023how2comm}. For instance, due to sensor malfunctions or electromagnetic interference (EMI), the data stream from a specific agent might be completely lost.
This leads to the loss of critical viewpoint information during data fusion, or even degrades the original overall perception performance.

To tackle these challenges, prior studies have introduced a series of robustness-oriented strategies, each crafted for a particular scenario. ERMVP \cite{zhang2024ermvp} mitigates spatial interference through consensus-driven sparse feature sampling. V2X-DG \cite{v2xdg} enhances fusion stability by enforcing cross-agent feature consistency. MDD \cite{mdd} integrates weather-invariant 4D radar to alleviate degradation caused by rain and snow. V2X-DGW \cite{v2xdgw} promotes generalization by simulating perception-range shrinkage and noise corruption during training. Although these methods show strong resilience to the distortions they target, their scalability remains limited. In complex real-world settings, interference arises from diverse and unpredictable sources, and models designed for a single corruption type often fail to adapt when new degradations appear. As illustrated in Fig.~\ref{fig:abstract}, their performance decreases sharply when facing unseen corruptions. For instance, MDD excels under water accumulation but declines markedly in the presence of EMI.

To address the aforementioned robustness bottleneck, we propose CoopDiff, a diffusion-guided \cite{diffusion} cooperative perception framework that explicitly models denoising objectives within the feature space. CoopDiff follows a teacher-student paradigm, in which the teacher generates high-quality denoised targets, while the student learns to reconstruct them through a generative denoising task.
The Quality-Aware Early-Fusion Teacher performs voxel-level early fusion by computing Quality of Interest (QoI) weights across multi-vehicle features and injecting semantic priors. The fused representation is then processed through a diffusion-based denoising module driven by Gated Conditional Modulation (GCM), producing a clean target feature map as supervision.
Meanwhile, the Dual-Branch Diffusion Student disentangles ego-centric and cooperative streams to model complementary denoising information. The local branch enhances the ego vehicle's environmental understanding, while the cooperative branch leverages the ego prior and employs Cooperative Deformable Attention (CDA) to adaptively sample and fuse informative features from other agents under degraded conditions. During decoding, Ego-Guided Cross-Attention (EGCA) enables interaction between local and cooperative representations, guiding the diffusion process to reconstruct denoised features from these two complementary perspectives.

To rigorously evaluate robustness under real-world degradations, we further construct two multi-corruption benchmarks, OPV2Vn and DAIR-V2Xn, which introduce six categories of corruption covering both environmental and sensor-level distortions. These datasets simulate realistic perception degradations and allow systematic testing of model resilience.
CoopDiff is trained on original data and evaluated under various corruption conditions. Extensive experiments on the OPV2Vn and DAIR-V2Xn benchmarks demonstrate that CoopDiff consistently surpasses existing state-of-the-art methods. It achieves substantially lower mean Relative Corruption Error (mRCE) across corruptions and exhibits minimal performance degradation in cross-domain tests, validating its strong generalization and robustness under distribution shifts and compound noise.

Our contributions are summarized as follows: (1) We present CoopDiff, a corruption-agnostic, diffusion-guided cooperative perception framework that explicitly tackles data corruptions and noise accumulation in multi-agent fusion. (2) CoopDiff integrates a Quality-Aware early-fusion Teacher with a Dual-Branch Diffusion Student, enabling adaptive feature optimization from both ego and cooperative views via QoI-weighted aggregation and guided denoising. (3) Experiments on OPV2Vn and DAIR-V2Xn demonstrate consistent SOTA performance and superior cross-domain generalization across diverse corruption scenarios.

\section{Related Work}

\noindent \textbf{Cooperative perception.}
Cooperative perception \cite{xu2022opv2v,dairv2x,coalign,v2xreal} enables vehicles to exchange information and cooperate across different fusion stages, typically categorized into early, intermediate, and late fusion. Among them, intermediate fusion has received the most attention due to its favorable balance between perception performance and communication bandwidth \cite{hu2022where2comm,fcooper,wang2020v2vnet}. Several milestone works have emerged in this domain. For instance, V2X-ViT \cite{xu2022v2xvit} employs a Transformer-based architecture to unify heterogeneous vehicle-infrastructure agents; Where2comm \cite{hu2022where2comm} optimizes the trade-off between accuracy and bandwidth by transmitting only spatially confident features; and CodeFilling \cite{codefilling} adopts a codebook-style representation to select information that best matches each agent's demand. 
However, most of these methods treat fusion as a structural problem operating on clean features, which provides optimal results under ideal conditions. In degraded viewpoints or noisy sensing conditions, feature errors tend to accumulate and become amplified through multi-agent fusion. To address this real-world gap, this work leverages a quality-aware early-fusion teacher to guide the student network in reconstructing noise-free features under noisy settings.

\noindent\textbf{Robust fusion.}
Recent research has begun to address degradation in real-world cooperative perception. ERMVP \cite{zhang2024ermvp} enhances collaboration in complex environments through consensus-driven feature sampling; the temporally aware MRCNet \cite{mrcnet} alleviates motion blur; and DSRC \cite{zhang2025dsrc} learns density-insensitive and semantic-aware cooperative features, yielding improvements on both clean and corrupted datasets. V2X-DGW \cite{v2xdgw} further mitigates performance deterioration in adverse weather by simulating degradation during training. Despite these advances, most existing methods are designed for specific perturbations or single-domain shifts, resulting in substantial performance drops when multiple corruptions or unseen domains are encountered. To fill this gap, we construct multi-degraded variants of OPV2V and DAIR-V2X to systematically evaluate model robustness under six individual corruption types.

\begin{figure*}[ht] 
  \centering 
  \includegraphics[width=0.98\linewidth]{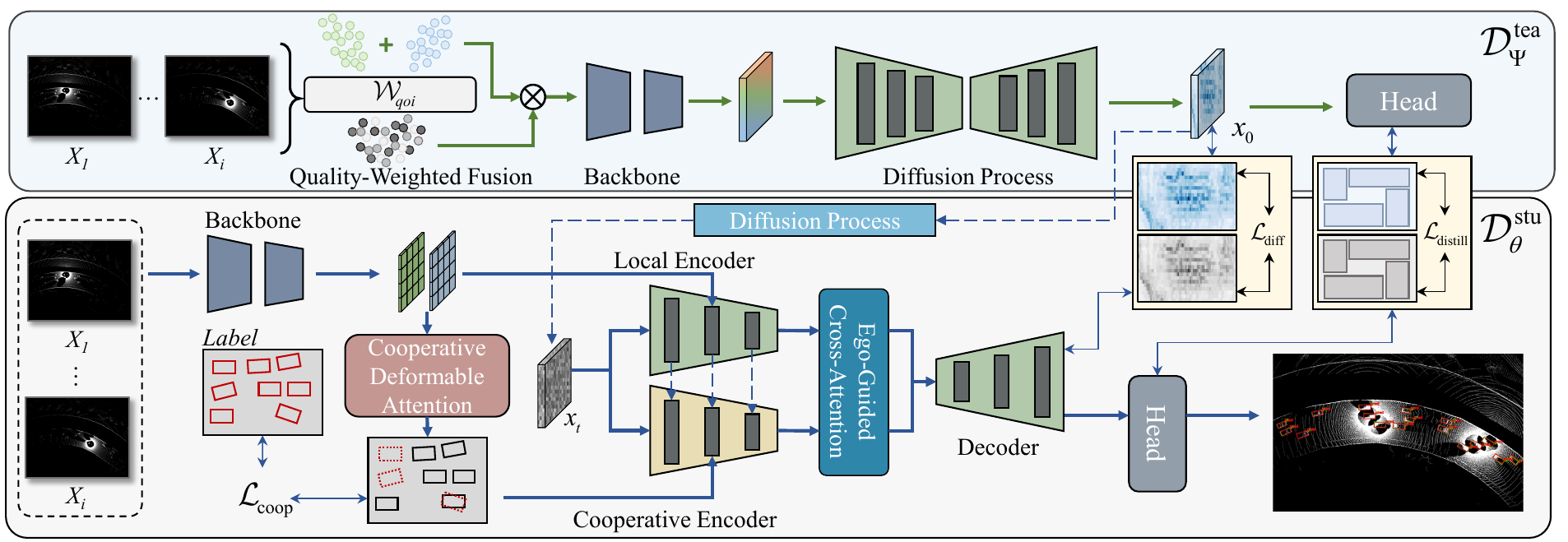}
  \caption{Overview of the proposed CoopDiff, which employs a Teacher-Student paradigm. The Quality-Aware Teacher model $\mathcal{D}_{\Psi}^{\text{tea}}$ uses an early-fusion strategy to process multi-agent inputs and generate a clean target feature map. The Dual-Branch Denoising Student $\mathcal{D}_{\theta}^{\text{stu}}$ is trained to reconstruct the target by leveraging local and cooperative conditions.}
  \label{fig:overall} 
\end{figure*}

\noindent \textbf{Diffusion methods.}
Diffusion models \cite{diffusion,diffusion2, diffusion3, ddim} have demonstrated remarkable denoising capability in 3D vision tasks, with representative examples including diff-pc \cite{diff-pc} and Mvdream \cite{Mvdream}. DifFUSER\cite{Diffuser} first applied diffusion to multimodal fusion by generating BEV features to handle noise; MDD \cite{mdd} leveraged diffusion for multimodal integration; and CoGMP \cite{CoGMP} incorporated generative map priors via diffusion to reduce transmission overhead. Building on these ideas, this work injects diffusion-driven denoising objectives directly into the feature space, enabling unified modeling of diverse and compound noise sources for multi-agent feature reconstruction and cooperative fusion.

\section{Method}
The goal of cooperative perception is to enable a group of $N$ connected agents $\mathcal{A} = \{a_1, \dots, a_N\}$, to achieve comprehensive scene understanding by sharing information. The system processes collective data $\mathcal{X} = \{X_j\}_{j=1}^N$, where $X_j$ represents the raw input from vehicle $a_j$, to generate a unified perception output $\mathcal{Y}$. 
A critical challenge arises from diverse data corruptions, which introduce noise and artifacts into the input $\mathcal{X}$. These corruptions severely degrade the quality of fusion and the accuracy of the output $\mathcal{Y}$.

To address this challenge, we propose CoopDiff, a diffusion-guided cooperative perception framework designed to enhance robustness against such corruptions. Fig. \ref{fig:overall} illustrates the architecture of CoopDiff. Our method employs a teacher-student paradigm featuring deep integration during the training phase. The teacher model $\mathcal{D}_{\Psi}^{\text{tea}}$ first performs quality-weighted fusion to suppress noisy inputs. Then, it applies a diffusion process to produce a denoised feature map. This map serves as the target $x_0$ for the student model $\mathcal{D}_\theta^{\text{stu}}$. The student model then executes its decoupled diffusion framework to construct the target $x_0$ by incorporating ego and cross-vehicle interaction information.

\begin{figure*}[ht] 
  \centering 
  \includegraphics[width=1.0\linewidth]{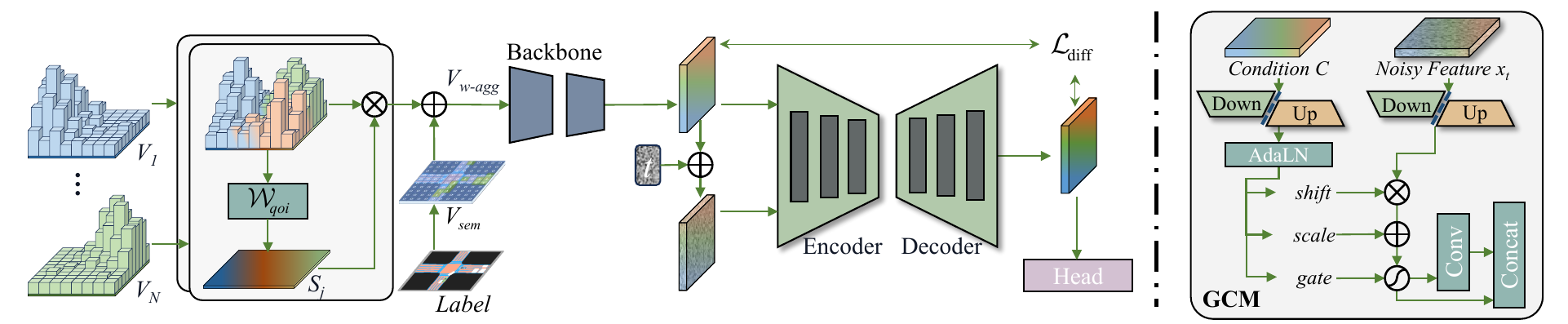}
  \caption{Overview of the Quality-Aware Early Fusion Teacher. Multi-agent features are first fused via Quality of Interest (QoI) weighting and semantic guidance, after which the GCM-based diffusion network denoises the input. The right shows the architecture of GCM.}
  \label{fig:teacher} 
\end{figure*}

\subsection{Quality-Aware Early Fusion Teacher}

While the early-fusion teacher paradigm is powerful \cite{DiscoNet, v2xdg}, it falters under data corruptions. Naive point cloud aggregation accumulates noise from all agents, yielding a low signal-to-noise ratio feature map that is an inefficient supervisor. To this end, we design a quality-aware early fusion teacher, as illustrated in Fig. \ref{fig:teacher}.

Specifically, we first generate dynamic QoI scores for each agent's voxelized features $V_j \in \mathbb{R}^{H \times W \times C_{in}}$. A shared convolutional module $\mathcal{W}_{qoi}$ estimates voxel-wise quality, producing a score map $S_j = \mathcal{W}_{qoi}(V_j)$, where $S_j \in \mathbb{R}^{H \times W \times 1}$. These scores act as QoI-driven weights to suppress corruption-prone regions during aggregation, thereby preserving stable geometric structures. The weighted aggregation $V_{w-agg}$ is:
\begin{equation}
V_{w-agg} = {\sum_{j=1}^{N} S_j \odot V_j}
\end{equation}
where $\odot$ denotes element-wise multiplication.

To enhance the teacher's representation from geometric to semantic levels, we further inject category information. Classification labels $L_{cls}$ are encoded into a dense semantic feature map $V_{sem} \in \mathbb{R}^{H \times W \times C_{sem}}$ via an embedding function $\mathcal{E}_{sem}$. This semantic prior is fused with the weighted geometric features via channel concatenation ($\mathbin{||}$).
\begin{equation}
V_{fused} = \text{Conv}(F_{w-agg} \mathbin{||} V_{sem})
\end{equation}

We adopt the standard DDPM \cite{diffusion}. Specifically, a backbone \(\mathcal{B}\) processes the fused representation \(V_{fused}\) to extract a conditional feature \(F^c = \mathcal{B}(V_{fused})\); we then sample a timestep \(t \sim \mathcal{U}(1, T)\) and apply the forward Q-process to the target \(F^c\) to obtain the noisy latent \(x_t\).
\begin{equation} \label{noise}x_t=\sqrt{\bar{\alpha}_t},F^c+\sqrt{1-\bar{\alpha}_t},\epsilon,\qquad \epsilon\sim\mathcal{N}(0,\mathbf{I}) 
\end{equation}

$F_c$ is then injected into our denoising diffusion network to guide its generative process. Specifically, $F_c$ is integrated with the noisy input $x_t$ at timestep $t$ via Gated Conditional Modulation (GCM) blocks. 
As shown in the right of Fig. \ref{fig:teacher}, our GCM is a FiLM-like \cite{perez2018film,Diffuser} operation: it passes the conditional feature through a small convolutional network to dynamically predict shift, scale, and gate parameters. These parameters then modulate the main diffusion backbone feature $x_{(l)}$, allowing the condition to precisely control the denoising trajectory at each layer.
Each layer $l$ of the network executes the following formula to predict the clean feature map \(x_0\).
\begin{equation}
x_{(l+1)}, F^c_{(l+1)} = \text{GCM}(F^c_{(l)}, x_{(l)}) + x_{(l)}
\end{equation}


After all layers, the final output is denoted as $x_0$, representing the fully denoised feature map reconstructed from the noisy input $x_t$. This clean feature $x_0$ serves as the supervision target for training our student network $\mathcal{D}_\theta^{\text{stu}}$.

\subsection{Dual-Branch Denoising Student}
\label{subsec:student}

Feature fusion under data corruption presents two challenges: retaining reliable ego features and extracting complementary information from collaborators.
However, a naive encoder that merges the noisy inputs before refinement tends to contaminate ego information with cooperative noise, thereby degrading its quality.
To address this, we reframe intermediate fusion as a generative denoising task. We propose a novel Dual-Branch Denoising Encoder that conditions the diffusion process to reconstruct the target $x_0$. It disentangles these two information sources, guiding the denoising process with local and cooperative perspectives.
Both conditional features are then injected into the decoder to guide the unified reconstruction.

\noindent \textbf{Local Encoder.} 
The student denoiser takes the noisy latent $x_t$ generated via $x_0$ like Eq.\ref{noise} and the timestep $t$ as its primary inputs.
The timestep $t$ is passed through a time-embedding layer to yield $\gamma(t)$. We then fuse $\gamma(t)$ with the ego-agent feature $F_i$ to form the initial, time-aware condition $c^{\text{loc}}_{(0)}=\phi_{\text{local}}(F_i)\;\oplus\;\gamma(t)$, where $\phi_{\text{local}}$ denotes a convolutional mapping and $\oplus$ denotes channel-wise addition.


The Local branch takes the noisy latent $x_t$ as its initial input. Let $x^{loc}_{(l)}$ be the feature map of this local backbone at level $l$ ($x^{loc}_{(0)} = x_t$), and $c_{(l)}^{loc}$ be the local condition. This branch consists of stacked GCM blocks that progressively refine the feature map by injecting the local condition:
\begin{equation}
x_{(l+1)}^{loc},c_{(l+1)}^{loc}=\text{GCM}^{(l)}(x_{(l)}^{loc},c_{(l)}^{loc})
\end{equation}
thereby guiding the refinement of the local feature stream.

\begin{figure}[t] 
  \centering 
  \includegraphics[width=0.95\linewidth]{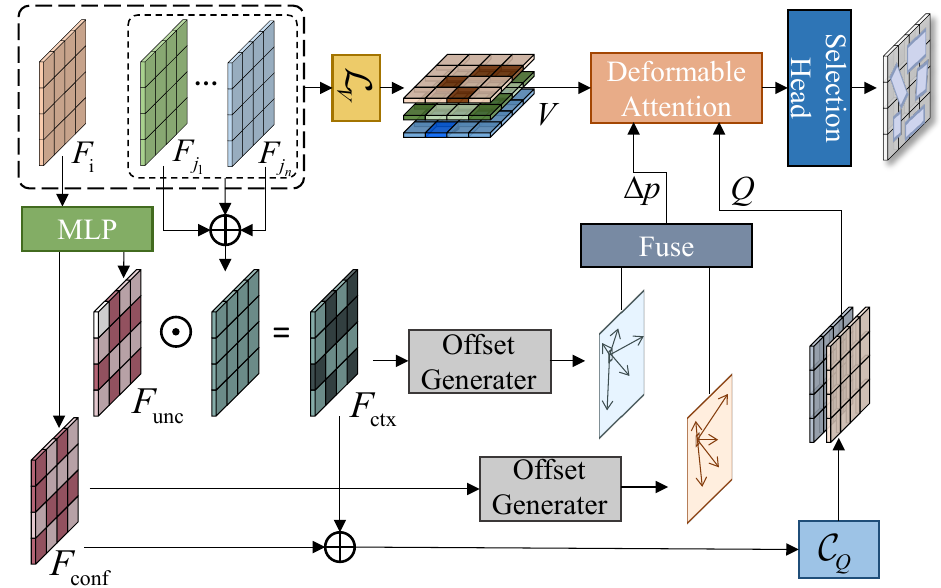}
  \caption{Architecture of the Cooperative Deformable Attention (CDA) module.}
  \label{fig:cda} 
\end{figure}

\noindent \textbf{Cooperative Encoder.} 
The Cooperative branch forms the multi-agent condition $c^{\text{coop}}$ using a Cooperative Deformable Attention (CDA) module.
First, the ego feature $F_i$ is enhanced via a MLP network \cite{mlp} and decoupled into high-confidence $F_{\text{conf}}$ and low-confidence $F_{\text{unc}}$ components. Subsequently, as depicted in Fig. \ref{fig:cda}, $F_{\text{unc}}$ is then concatenated with features from collaborating agents to construct the cooperative context:
\begin{equation}
F_{\text{ctx}}=\text{Agg}\;\!\big(\{\,\phi_{\text{coop}}(F_j)\,\}_{j\neq i}\big)\;\odot\;F_{\text{unc}},
\end{equation}
where $\phi_{\text{coop}}$ is a feature mapping, $\text{Agg}$ aligns channels via convolution, and $\odot$ is the element-wise multiplication.

Individual offsets are predicted from $F_{\text{conf}}$ and $F_{\text{ctx}}$ separately to capture position adjustments from the ego and cooperative contexts.
These two offsets are fused through a convolutional layer to form the final offset $\Delta p$, which is used by the deformable attention \cite{deformatt} for precise sampling.
\begin{equation}
F_{\text{coop}} = \text{DeformAttn}\big(Q,\,V,\,\Delta p\big), 
\end{equation}
\begin{equation}
Q = \mathcal{C}_Q\big(F_{\text{conf}}\mathbin{||} F_{\text{ctx}}\big), \quad 
V = \mathcal{L}_V(F_{\text{ctx}}),
\end{equation}
where $\mathcal{C}_Q$ is a $1 \times 1$ convolution, $\mathcal{L}_V$ is a linear projection. 

To optimize the fused representation, $F_{coop}$ is then processed by a selection head to retain only the most informative tokens, yielding a sparse feature map $\tilde F_{coop}$.
At each pyramid level $l$, $\tilde F_{\text{coop}}$ is concatenated with the local prior $c^{\text{loc}}_{(l)}$ via a convolution to form the cooperative condition $c^{\text{coop}}_{(l)}$.
This encoder is also modulated by GCM at every level $l$ and takes the latent $x_t$ as its initial input $x^{coop}_{(0)}$.
\begin{equation}
x^{coop}_{(l+1)},\;c^{\text{coop}}_{(l+1)}=\text{GCM}^{(l)}\!\big(x^{coop}_{(l)},\,c^{\text{coop}}_{(l)}\big)
\end{equation}

This design allows the denoising process to simultaneously preserve high-confidence ego features and integrate complementary cross-agent information.

\noindent \textbf{Decoder.} 
The refined outputs from the parallel encoder branches, $x^{\text{loc}}$ and $x^{\text{coop}}$, are then fused within the upsampling decoder for unified reconstruction. This design allows the decoder to balance ego-feature integrity with complementary cooperative information using our proposed Ego-Guided Cross-Attention (EGCA) module.

Specifically, We project the query $Q$ exclusively from $x^{\text{loc}}$, while the key-value pairs are sourced from both branches: $(K_{loc}, V_{loc})$ from $x^{\text{loc}}$ and $(K_{coop}, V_{coop})$ from $x^{\text{coop}}$. A shared positional encoding $P$ is then added to $Q$, $K_{loc}$, and $K_{coop}$ to maintain spatial correspondence.

The unified key $K$ and value $V$ are constructed as $K = [K_{loc} \Vert K_{coop}]$ and $V = [V_{loc} \Vert V_{coop}]$, where $\Vert$ denotes sequence concatenation. The attention output $F_{att}$ is computed via standard cross attention \cite{attention}:
\begin{equation}
F_{att} = \text{CrossAttn}(Q, K, V).
\end{equation}

Finally, the representation $F_{att}$ is passed to the prediction head to generate the perception output $\mathcal{Y}$.

\subsection{Loss}


The student $\mathcal{D}_\theta^{\text{stu}}$ is trained to master two objectives: reconstructing the clean target $x_0$ and the downstream perception task. The core detection loss, $\mathcal{L}_{\text{task}}$, is a standard composite of classification \cite{lin2017focal} and regression \cite{wu2018l1} objectives. The full objective $\mathcal{L}_{\text{total}}$ adds three further constraints: diffusion ($\mathcal{L}_{\text{diff}}$), knowledge distillation ($\mathcal{L}_{\text{distill}}$), and cooperative supervision ($\mathcal{L}_{\text{coop}}$). These four components are balanced by hyperparameters $\alpha, \beta, \gamma, \delta$.
\begin{equation}
\mathcal{L}_{\text{total}} = \alpha \mathcal{L}_{\text{task}} + \beta \mathcal{L}_{\text{diff}} + \gamma \mathcal{L}_{\text{distill}} + \delta \mathcal{L}_{\text{coop}}
\end{equation}


\noindent \textbf{Diffusion Loss ($\mathcal{L}_{\text{diff}}$)}. 
The student $\mathcal{D}_\theta^{\text{stu}}$ is trained to predict the noise $\hat y_t$ based on the noisy input $x_t$, timestep $t$, and conditions ($c^{\text{loc}}, c^{\text{coop}}$), targeting the ground-truth noise $y_t = \epsilon$. 
We further employ a heteroscedastic NLL \cite{nll} to enhance robustness, where the student model predicts a log-variance $s_t=\log\sigma_t^2$ to quantify its noise prediction uncertainty.
\begin{equation}
\mathcal{L}_{\text{diff}}
=\mathbb{E}_{t,\epsilon}\left[\frac{1}{2}e^{-s_t}|y_t-\hat y_t|_2^2+\frac{1}{2}s_t\right]
\label{eq:diff_unc}
\end{equation}

This formulation down-weights inherently ambiguous pixels and enhances numerical stability.

\noindent \textbf{Knowledge Distillation Loss ($\mathcal{L}_{\text{distill}}$)}.
To enhance semantic transfer at the decision level, we introduce an additional logits-based distillation term $\mathcal{L}_{\text{distill}}$. This complements the feature-level supervision already provided by the $\mathcal{L}_{\text{diff}}$ objective, which aligns the student's denoised representation with the teacher's target $x_0$. We align the student's output logits ($z^{\text{stu}}$) with the teacher's ($z^{\text{tea}}$) by minimizing the Kullback-Leibler (KL) divergence \cite{kl}:
\begin{equation}
\mathcal{L}_{\text{distill}} = \mathcal{D}_{\text{KL}}\big( \sigma(z^{\text{tea}} / \tau) || \sigma(z^{\text{stu}} / \tau) \big),
\end{equation}
where $\sigma$ denotes the softmax function, and $\tau=1$ is the distillation temperature. This term complements the feature-level supervision by transferring the teacher's decision boundaries, encouraging corruption-robust and semantically consistent predictions.

\noindent \textbf{Cooperative Supervision Loss ($\mathcal{L}_{\text{coop}}$)}.
To guide the cooperative branch, we apply a supervision loss $\mathcal{L}_{\text{coop}}$ to the spatial selection map $M_{coop}$ generated by its selection head (as mentioned in Sec.~\ref{subsec:student}). This map $M_{coop}$ is trained to identify regions where collaboration is necessary by predicting ground-truth cooperative masks $Y_{\text{gt}}$. We adopt a binary cross-entropy (BCE) \cite{bce} objective:
\begin{equation}
\mathcal{L}_{\text{coop}} = \mathcal{L}_{\text{BCE}}(M_{coop}, Y_{\text{gt}}),
\end{equation}
which drives the selection head to activate on necessary foreground regions.

\begin{table*}[tbp]
    \centering
    \caption{Performance of different models on the OPV2V dataset under various conditions.}
    \label{tab:opv2v_results}
    \begin{adjustbox}{max width=\linewidth}
    \begin{tabular}{c *{14}{S[table-format=1.4]}}
    \hline
\multirow{2}{*}{OPV2Vn} & \multicolumn{2}{c}{Clean Data} & \multicolumn{2}{c}{Beam Missing} & \multicolumn{2}{c}{Motion Blur} & \multicolumn{2}{c}{Fog} & \multicolumn{2}{c}{EMI} & \multicolumn{2}{c}{Water} & \multicolumn{2}{c}{Echo} \\
    \cmidrule(lr){2-3} \cmidrule(lr){4-5} \cmidrule(lr){6-7} \cmidrule(lr){8-9} \cmidrule(lr){10-11} \cmidrule(lr){12-13} \cmidrule(lr){14-15}
    & {AP 0.5} & {AP 0.7} & {AP 0.5} & {AP 0.7} & {AP 0.5} & {AP 0.7} & {AP 0.5} & {AP 0.7} & {AP 0.5} & {AP 0.7} & {AP 0.5} & {AP 0.7} & {AP 0.5} & {AP 0.7} \\
    \hline
Single & 0.7381 & 0.5081 & 0.5591 & 0.3289 & 0.6050 & 0.2798 & 0.5378 & 0.4014 & 0.6046 & 0.3687 & 0.7197 & 0.4612 & 0.6701 & 0.4612 \\
Intermediate & 0.7223 & 0.5699 & 0.5411 & 0.3623 & 0.6139 & 0.3672 & 0.5164 & 0.4349 & 0.5682 & 0.3832 & 0.7026 & 0.5284 & 0.6773 & 0.5325 \\
Late & 0.8484 & 0.7694 & 0.7574 & 0.6204 & 0.7254 & 0.4781 & 0.6798 & 0.6152 & 0.7552 & 0.6239 & 0.8413 & 0.7580 & 0.8156 & 0.7382 \\
Fcooper \cite{fcooper} & 0.8800 & 0.7852 & 0.7460 & 0.6122 & 0.7121 & 0.4727 & 0.6440 & 0.5687 & 0.7563 & 0.6094 & 0.8676 & 0.7642 & 0.8519 & 0.7573 \\
V2VNet \cite{wang2020v2vnet} & 0.8840 & 0.7613 & 0.7450 & 0.5681 & 0.7963 & 0.5522 & 0.6223 & 0.5091 & 0.7630 & 0.5794 & 0.8617 & 0.7117 & 0.8550 & 0.7313 \\
 V2X-ViT \cite{xu2022v2xvit} & 0.8740 & 0.7692 & 0.7421 & 0.5960 & 0.7101 & 0.4723 & 0.6184 & 0.5351 & 0.7647 & 0.6010 & 0.8496 & 0.7328 & 0.8586 & 0.7522 \\
Where2Comm \cite{hu2022where2comm} & 0.8757 & 0.7451 & 0.7567 & 0.5767 & 0.7823 & 0.5066 & 0.6702 & 0.5654 & 0.7470 & 0.5729 & 0.8614 & 0.7137 & 0.8598 & 0.7259 \\
Coalign \cite{coalign} & 0.8878 & 0.7931 & 0.7657 & 0.6308 & 0.7785 & 0.5025 & 0.6778 & 0.6016 & 0.7741 & 0.6374 & 0.8690 & 0.7598 & 0.8642 & 0.7662 \\
ERMVP \cite{zhang2024ermvp} & 0.8398 & 0.7211 & 0.6684 & 0.5225 & 0.6763 & 0.4365 & 0.5407 & 0.4613 & 0.6202 & 0.4645 & 0.7526 & 0.6102 & 0.8221 & 0.6997 \\
MRCNet \cite{mrcnet} & 0.8529 & 0.7699 & 0.7190 & 0.6301 & 0.7107 & 0.5402 & 0.6424 & 0.5969 & 0.6891 & 0.6017 & 0.8050 & 0.6862 & 0.8545 & 0.7723 \\
DSRC \cite{zhang2025dsrc} & 0.8941 & 0.8035 & 0.7894 & 0.6648 & 0.8062 & 0.5633 & 0.6763 & 0.6119 & 0.7714 & 0.6394 & 0.8649 & 0.7520 & 0.8816 & 0.7909 \\
MDD \cite{mdd} & 0.7792 & 0.6356 & 0.6047 & 0.4513 & 0.6566 & 0.3566 & 0.5375 & 0.4472 & 0.5645 & 0.4261 & 0.7513 & 0.5949 & 0.6475 & 0.4961 \\
\hline
Ours & 0.9053 & 0.8357 & 0.7909 & 0.7020 & 0.8142 & 0.6184 & 0.6871 & 0.6297 & 0.7891 & 0.6897 & 0.8745 & 0.7963 & 0.8923 & 0.8200 \\
    \hline
    \end{tabular}
    \end{adjustbox}
\end{table*}

\begin{table*}[tbp]
    \centering
    \caption{Performance of different models on the DAIR-V2X dataset under various conditions.}
    \label{tab:dairv2x_results}
    \begin{adjustbox}{max width=\linewidth}
    \begin{tabular}{c *{14}{S[table-format=1.4]}}
    \hline
\multirow{2}{*}{DAIR-V2Xn} & \multicolumn{2}{c}{Clean Data} & \multicolumn{2}{c}{Beam Missing} & \multicolumn{2}{c}{Motion Blur} & \multicolumn{2}{c}{Fog} & \multicolumn{2}{c}{EMI} & \multicolumn{2}{c}{Water} & \multicolumn{2}{c}{Echo} \\
    \cmidrule(lr){2-3} \cmidrule(lr){4-5} \cmidrule(lr){6-7} \cmidrule(lr){8-9} \cmidrule(lr){10-11} \cmidrule(lr){12-13} \cmidrule(lr){14-15}
    & {AP 0.5} & {AP 0.7} & {AP 0.5} & {AP 0.7} & {AP 0.5} & {AP 0.7} & {AP 0.5} & {AP 0.7} & {AP 0.5} & {AP 0.7} & {AP 0.5} & {AP 0.7} & {AP 0.5} & {AP 0.7} \\
    \hline
Single & 0.6630 & 0.5152 & 0.3370 & 0.2598 & 0.5883 & 0.3713 & 0.3486 & 0.2557 & 0.3459 & 0.2629 & 0.5976 & 0.4966 & 0.6377 & 0.5024 \\
Intermediate & 0.6973 & 0.5111 & 0.3591 & 0.2305 & 0.6312 & 0.3653 & 0.3579 & 0.2507 & 0.3545 & 0.2483 & 0.6373 & 0.4291 & 0.6917 & 0.5012 \\
Late & 0.6505 & 0.5538 & 0.3322 & 0.2080 & 0.5423 & 0.3437 & 0.3777 & 0.2920 & 0.3713 & 0.2498 & 0.5827 & 0.4114 & 0.6107 & 0.4350 \\
Fcooper \cite{fcooper} & 0.7396 & 0.5612 & 0.3871 & 0.2628 & 0.6066 & 0.3547 & 0.3696 & 0.2775 & 0.3801 & 0.2766 & 0.6738 & 0.4916 & 0.7299 & 0.5539 \\
V2VNet \cite{wang2020v2vnet} & 0.6392 & 0.3132 & 0.3582 & 0.1518 & 0.4533 & 0.1892 & 0.3144 & 0.1528 & 0.3422 & 0.1408 & 0.5800 & 0.2570 & 0.6264 & 0.3039 \\
 V2X-ViT \cite{xu2022v2xvit} & 0.7156 & 0.5464 & 0.3685 & 0.2400 & 0.6195 & 0.3829 & 0.3967 & 0.3037 & 0.3800 & 0.2599 & 0.6401 & 0.4616 & 0.6938 & 0.5193 \\
Where2Comm \cite{hu2022where2comm} & 0.6737 & 0.5319 & 0.3195 & 0.2477 & 0.5568 & 0.3823 & 0.3801 & 0.3067 & 0.3542 & 0.2699 & 0.6215 & 0.4731 & 0.6665 & 0.5244 \\
Coalign \cite{coalign} & 0.7773 & 0.6284 & 0.4901 & 0.3395 & 0.6771 & 0.4559 & 0.4295 & 0.3528 & 0.4577 & 0.3415 & 0.7143 & 0.5626 & 0.7721 & 0.6190 \\
ERMVP \cite{zhang2024ermvp} & 0.6738 & 0.5546 & 0.3216 & 0.2599 & 0.5616 & 0.4027 & 0.3837 & 0.3202 & 0.3535 & 0.2836 & 0.6217 & 0.4920 & 0.6680 & 0.5470 \\
MRCNet \cite{mrcnet} & 0.6649 & 0.5388 & 0.3718 & 0.2653 & 0.6054 & 0.4204 & 0.3704 & 0.3118 & 0.3587 & 0.2757 & 0.5863 & 0.4578 & 0.6636 & 0.5373 \\
DSRC \cite{zhang2025dsrc} & 0.7517 & 0.6164 & 0.4225 & 0.3117 & 0.6452 & 0.4453 & 0.4152 & 0.3478 & 0.4211 & 0.3184 & 0.6919 & 0.5453 & 0.7406 & 0.6061 \\
MDD \cite{mdd} & 0.7507 & 0.5813 & 0.4448 & 0.3021 & 0.6647 & 0.4493 & 0.4179 & 0.3234 & 0.3750 & 0.2615 & 0.6702 & 0.5044 & 0.7321 & 0.5614 \\
\hline
Ours & 0.8069 & 0.6644 & 0.5243 & 0.3597 & 0.6830 & 0.4690 & 0.4359 & 0.3623 & 0.4734 & 0.3555 & 0.7380 & 0.5834 & 0.7972 & 0.6562 \\
    \hline
    \end{tabular}
    \end{adjustbox}
\end{table*}

\begin{figure*}[t] 
    \centering
    \begin{subfigure}[t]{0.47\textwidth} 
        \centering
        \includegraphics[width=\linewidth]{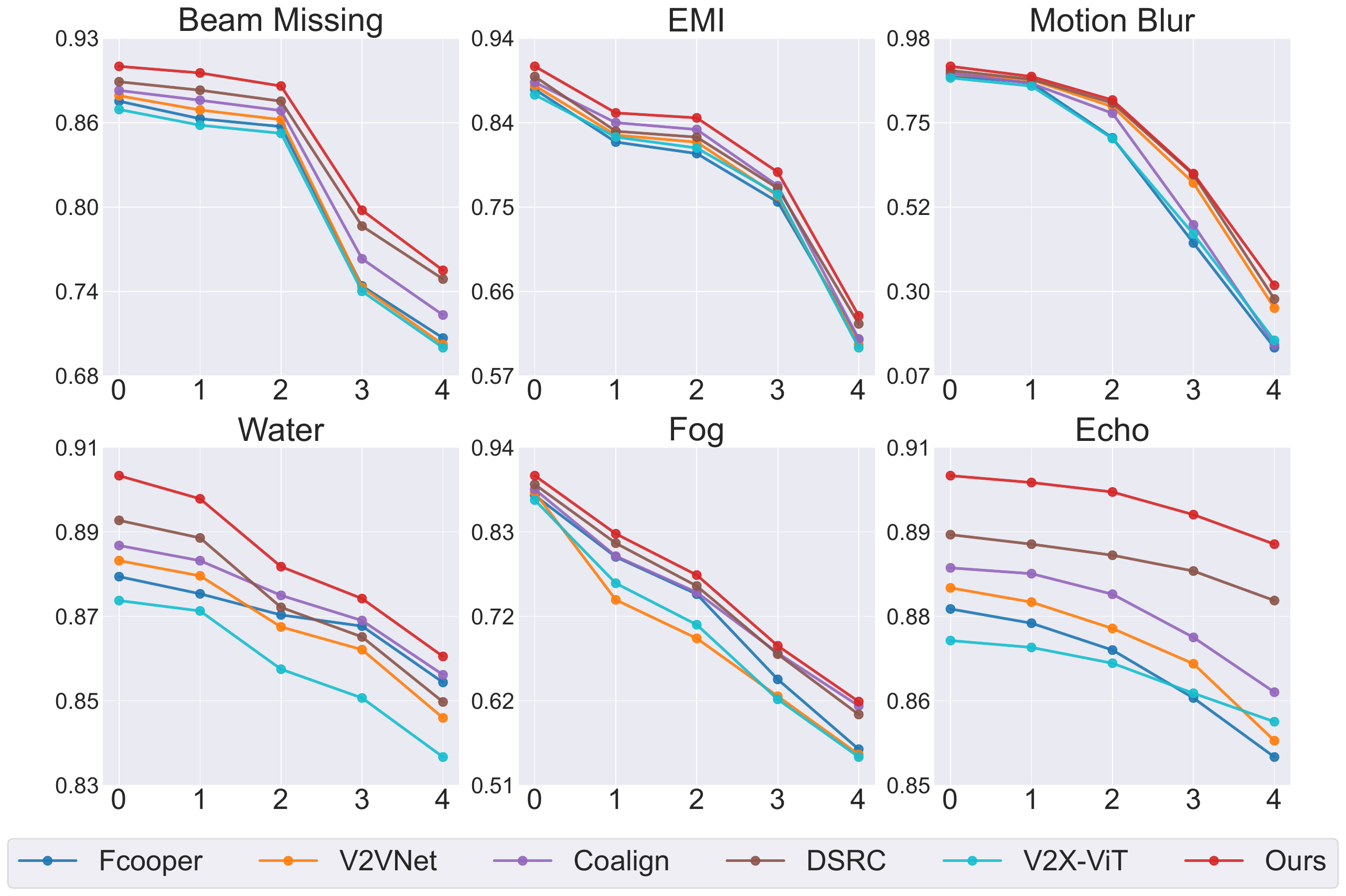}
        \caption{Comparison on OPV2Vn.} 
        \label{fig:sub_opv2v}
    \end{subfigure}
    \hfill 
    \begin{subfigure}[t]{0.47\textwidth} 
        \centering
        \includegraphics[width=\linewidth]{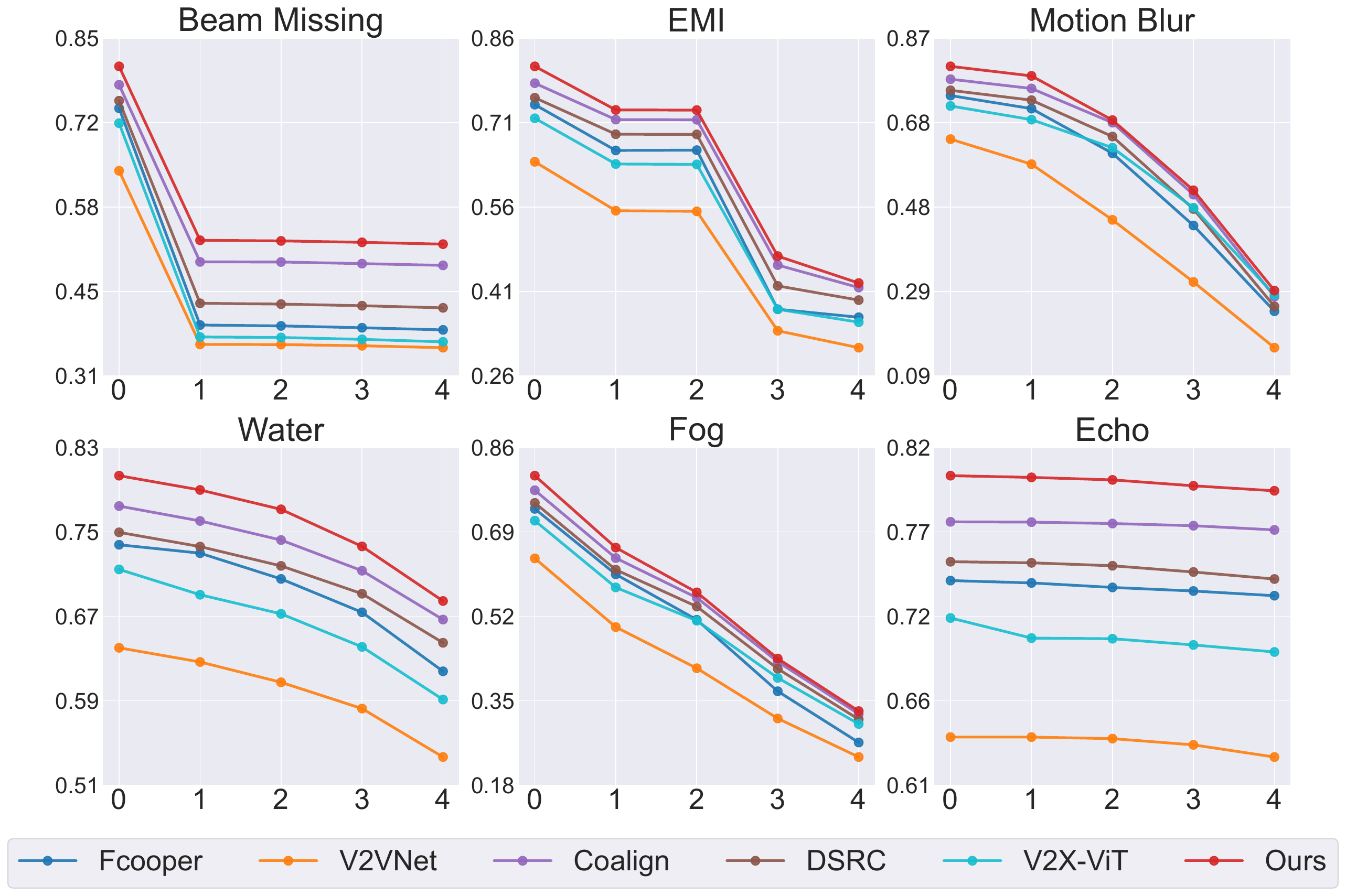}
        \caption{Comparison on DAIR-V2Xn.} 
        \label{fig:sub_dair}
    \end{subfigure}
    \vspace{-3mm} 
    \caption{Robustness of our proposed method and compared benchmark to varying levels of corruption.} 
    \label{fig:robustness} 
\end{figure*}

\section{Experiments}

\subsection{Datasets and Experimental Settings}

\noindent \textbf{Datasets.} To validate the model's effectiveness under severe corruptions, we considered the impact of multiple corruption scenarios. We systematically augment the original datasets with six different types of corruption to simulate a wide range of real-world challenges. These corruptions are organized into two main categories. 1) \textit{External Environmental Corruptions}, such as the effects of fog, rain and standing water, and electromagnetic interference. 2) \textit{Internal Sensor Artifacts}, such as motion blur, multi-path reflections, and beam missing. Based on the widely-used datasets DAIR-V2X \cite{dairv2x} and OPV2V \cite{xu2022opv2v}, we created two point cloud corruptions benchmarks, DAIR-V2Xn and OPV2Vn, which are built upon Robo3D \cite{kong2023robo3d}, to evaluate the resilience against environmental and sensor level corruptions.

\noindent \textbf{Implementation Details.} 
All the compared methods were retrained based on the OpenCOOD \cite{xu2022opv2v} framework. We adopted PointPillars \cite{lang2019pointpillars} as the LiDAR encoder
.
During training, all models were trained on clean point cloud data for 30 epochs with batch size of 2 to evaluate generalization ability. The Adam optimizer \cite{kingma2014adam} was used with an initial learning rate of 1e-3, and the number of cooperating vehicles was set to 2.
In our method, both the diffusion encoder and decoder consist of 3 layers, and the selection ratio is set to 30\%. 
We train the model using the DDPM\cite{diffusion} and employ the deterministic DDIM\cite{ddim} during inference.
Intermediate features are compressed by 16$\times$, and the teacher model is utilized exclusively during training.


\subsection{Quantitative evaluation}

\noindent \textbf{Benchmark Comparison.} Tab.~\ref{tab:opv2v_results} and Tab.~\ref{tab:dairv2x_results} present a comprehensive comparison between our proposed method and leading baselines across the OPV2Vn and DAIR-V2Xn, evaluated under both clean and corrupted conditions. Across all scenarios, our method consistently surpassing the current SOTA methods.
\textit{(1)} Under clean conditions, CoopDiff achieve performance gains of 1.12\% / 3.22\% and 2.96\% / 3.60\% in AP@0.5/AP@0.7 over the second-best method on OPV2Vn and DAIR-V2Xn. This confirms that the proposed diffusion-driven framework produces more stable feature representations, even when processing high-fidelity clean data.
\textit{(2)} in corrupted conditions, CoopDiff shows remarkable robustness. On average, across all six corruption types, CoopDiff outperforms the baseline average by 8.40\% / 13.16\% on OPV2Vn and 10.24\% / 10.13\% on DAIR-V2Xn. 
This massive lead highlights its ability to maintain high performance across diverse corruption scenarios. Notably, MDD exhibits clear performance degradation when limited to LiDAR-only inputs, highlighting the limitations of existing methods. This advantage stems from our diffusion-guided denoising paradigm and the dual-branch encoder, which jointly enable adaptive noise suppression and cross-agent information fusion.

\noindent \textbf{Corruption Sensitivity Analysis.} As shown in Fig.~\ref{fig:robustness}, we systematically evaluate the sensitivity of existing cooperative perception models on OPV2Vn and DAIR-V2Xn at varying severity levels.
All baselines exhibit severe degradation under motion blur and fog conditions. 
For instance, under the strongest blur, Coalign on OPV2Vn drops from 0.8878 to 0.1529, while DSRC on DAIR-V2X falls from 0.7517 to 0.2540. 
In contrast, CoopDiff consistently surpasses all prior SOTA models across all corruption types and severity levels, demonstrating the lowest sensitivity.

\begin{table}[t]
\small
\centering
\caption{RCE comparison where mRCE is the average of RCE@0.5 and 0.7. The format is RCE@0.5 / RCE@0.7 / mRCE.}
\label{tab:mRCE}
\begin{adjustbox}{max width=\linewidth}
\begin{tabular}{lcc}
\hline
Model & OPV2V (\% $\downarrow$) & DAIR-V2X (\% $\downarrow$) \\
\hline
Fcooper      & 13.06 / 25.13 / 19.10 & 33.15 / 33.33 / 33.24 \\
V2VNet       & 10.89 / 26.69 / 18.79 & 46.54 / 50.84 / 48.69 \\
 V2X-ViT      & 12.30 / 26.61 / 19.46 & 33.05 / 33.30 / 33.18 \\
Coalign      & 11.95 / 23.36 / 17.66 & 28.89 / 28.91 / 28.90 \\
ERMVP        & 19.38 / 31.91 / 25.65 & 29.35 / 29.87 / 29.61 \\
MRCNet       & 13.78 / 21.04 / 17.41 & 30.73 / 30.56 / 30.65 \\
DSRC         & 11.13 / 20.40 / 15.77 & 30.45 / 30.63 / 30.54 \\
MDD          & 19.00 / 31.06 / 25.03 & 32.61 / 33.04 / 32.83 \\
\hline
Ours         & 10.75 / 15.12 / 12.94 & 26.73 / 26.85 / 26.79 \\
\hline
\end{tabular}
\end{adjustbox}
\end{table}

\begin{table}[t]
  \centering
  \caption{Performance Comparison of Cooperative Perception Methods across Different Scenarios.}
  \label{tab:domain_gen}
  \resizebox{\columnwidth}{!}{
    \begin{tabular}{lcccc}
      \hline
      \multirow{2}{*}{Model} & \multicolumn{2}{c}{AP@0.5} & \multicolumn{2}{c}{AP@0.7} \\
      \cline{2-5}
       & \multicolumn{1}{c}{Default} & \multicolumn{1}{c}{Culver} & \multicolumn{1}{c}{Default} & \multicolumn{1}{c}{Culver} \\
      \hline
      V2VNet & 0.8840 & 0.8424 (04.71\% $\downarrow$) & 0.7613 & 0.6924 (09.05\% $\downarrow$) \\
       V2X-ViT & 0.8740 & 0.8328 (04.71\% $\downarrow$) & 0.7692 & 0.6884 (10.50\% $\downarrow$) \\
      Coalign & 0.8878 & 0.8418 (05.18\% $\downarrow$) & 0.7931 & 0.7121 (10.21\% $\downarrow$) \\
      MRCNet & 0.8529 & 0.7444 (12.72\% $\downarrow$) & 0.7699 & 0.6364 (17.34\% $\downarrow$) \\
      DSRC & 0.8941 & 0.8311 (07.05\% $\downarrow$) & 0.8035 & 0.7076 (11.94\% $\downarrow$) \\
      Ours & 0.9053 & 0.8641 (04.55\% $\downarrow$) & 0.8357 & 0.7563 (09.50\% $\downarrow$) \\
      \hline
    \end{tabular}%
  }
\end{table}%

\noindent \textbf{Quantitative Robustness Analysis.} Tab.~\ref{tab:mRCE} further reports the average relative performance drop across corruptions. 
CoopDiff achieves the lowest mRCE of 12.94\% on OPV2V and 26.79\% on DAIR-V2X, quantitatively confirming its superior resilience against diverse degradations. 
We attribute this resilience to two key factors: 
(1) the student network effectively inherits the teacher's \emph{quality-aware supervision}, suppressing noise accumulation during feature denoising; and 
(2) the dual-branch local/cooperative feature interaction mitigates performance decay caused by corrupted collaboration, maintaining stable detection performance even under severe interference.

\noindent \textbf{Generalization to Unseen Domains.} Tab.~\ref{tab:domain_gen} shows the models' generalization capability when tested on the unseen Culver City domain after being trained only on the default OPV2V dataset. Our approach achieves the highest AP scores and suffers the smallest performance drop (only 4.55\% in AP@0.5), demonstrating superior domain generalization compared to all baselines.

\begin{table}[t]
  \centering
  \caption{Ablation study of our proposed modules.}
  \label{tab:performance_comparison_simplified}
  \begin{adjustbox}{max width=\linewidth}
    \begin{tabular}{cccccc}
      \hline
      GCM & Coop & Teach & EGCA & OPV2V & DAIR-V2X \\
      \hline
      & & & & 0.8548 / 0.7431 & 0.7330 / 0.5530 \\
      \checkmark & & & & 0.8681 / 0.7765 & 0.7565 / 0.6092 \\
      & \checkmark & & & 0.8705 / 0.8014 & 0.7918 / 0.6361 \\ 
      \checkmark & & \checkmark & & 0.8862 / 0.8192 & 0.8027 / 0.6591 \\
      \checkmark & \checkmark & & & 0.8852 / 0.8162 & 0.7971 / 0.6517 \\
      \checkmark & \checkmark & \checkmark & & 0.8984 / 0.8294 & 0.8042 / 0.6610 \\
      \checkmark & \checkmark & \checkmark & \checkmark & 0.9053 / 0.8357 & 0.8069 / 0.6644 \\
      \hline
    \end{tabular}%
  \end{adjustbox}
\end{table}%

\subsection{Component Analysis}

\noindent \textbf{Contribution of Major Components.} Tab.~\ref{tab:performance_comparison_simplified} summarizes the contribution of each proposed component. 
Integrating the the GCM-driven diffusion mechanism into baseline yields 1.33\%/3.34\% gains on OPV2V and 2.35\%/5.62\% gains on DAIR-V2X. This confirms the diffusion-based mechanism enhances local feature representations. The Cooperative (Coop) branch further boosts performance (e.g., +5.3\% AP@0.5 on DAIR-V2X), indicating cross-agent fusion provides complementary spatial cues. The Teacher (Teach) branch adds a stable gain, suggesting its features help suppress noise accumulation. Finally, the full model equipped with EGCA achieves the optimal accuracy: 0.9053/0.8357 on OPV2V and 0.8069/0.6644 on DAIR-V2X. This validation confirms that each component is indispensable for achieving the optimal performance.


\noindent \textbf{Ablation on Teacher Model.} Tab.~\ref{tab:teacher_model_comparison_combined} validates our teacher model design. Compared to a Naive Early baseline, incorporating QoI weighting yields a performance boost, highlighting the importance of suppressing interference during aggregation. Integrating the Semantic Guidance Mask (SGM) to inject category-level priors provides an additional gain. 
The teacher integrated with diffusion can better learn the feature distribution of the generation process, thereby providing more effective guidance for the student.

\noindent \textbf{Number of Diffusion Steps.} Tab.~\ref{tab:performance_comparison_revised} analyzes the framework's adjustable trade-off between accuracy and efficiency. Notably, our model is highly efficient, surpassing the previous state-of-the-art performance with only 2 diffusion steps while operating at 9.45 FPS on OPV2V. Increasing the steps to 10 further refines geometric details and boosts accuracy. However, gains with more steps become negligible ($<$0.1\%). We use 10 steps to report its accuracy.

\begin{table}[t]
\centering
\caption{Ablation study on Quality-Aware Early Fusion Teacher.}
\label{tab:teacher_model_comparison_combined}
\begin{adjustbox}{max width=\linewidth}
\begin{tabular}{lcc}
\hline
Teacher Model & OPV2V & DAIR-V2X \\
\hline
Naive Early & 0.8775 / 0.7717 & 0.7756 / 0.6011 \\
+ QoI    & 0.8927 / 0.8258 & 0.7976 / 0.6404 \\
+ QoI + SGM    & 0.9048 / 0.8325 & 0.8061 / 0.6533 \\
+ QoI + SGM + GCM &  0.9053 / 0.8357 & 0.8069 / 0.6644 \\
\hline
\end{tabular}
\end{adjustbox}
\end{table}


\begin{table}[t]
  \centering
  \caption{Ablation study on the number of diffusion steps.}
  \label{tab:performance_comparison_revised}
  \resizebox{\columnwidth }{!}{  
    \begin{tabular}{ccccc}
      \hline
      Number & OPV2V & FPS & DAIR-V2X & FPS \\
      \hline
      1 & 0.8602/0.8152 & 10.14 & 0.7753/0.6309 & 9.47 \\
      2 & 0.8970/0.8326 & 9.45 & 0.8000/0.6512 & 8.19 \\
      5 & 0.9046/0.8345 & 7.33 & 0.8028/0.6594 & 7.02 \\
      10 & 0.9053/0.8357 & 5.31 & 0.8069/0.6644 & 4.89 \\
      20 & 0.9060/0.8362 & 3.22 & 0.8073/0.6650 & 3.14 \\
      \hline
    \end{tabular}%
  }
\end{table}

\begin{table}[t]
  \centering
  \caption{Performance with Varying Local Data Selection Ratios}
  \label{tab:local_ratio_ablation}
  \begin{adjustbox}{max width=\textwidth}
    \begin{tabular}{ccc}
      \hline
      Ratio & OPV2V & DAIR-V2X \\
      \hline
      50\% & 0.9051/0.8354 & 0.8069/0.6642 \\
      30\% & 0.9053/0.8357 & 0.8069/0.6644 \\
      20\% & 0.9046/0.8349 & 0.8061/0.6380 \\
      10\% & 0.9040/0.8339 & 0.8059/0.6375 \\
      5\% & 0.9031/0.8325 & 0.8054/0.6344 \\
      \hline
    \end{tabular}%
  \end{adjustbox}
\end{table}%

\noindent \textbf{Ablation on Selection Ratio.} Tab.~\ref{tab:local_ratio_ablation} investigates the impact of the Top-K selection ratio in our cooperative branch. Performance on both datasets remains stable. This confirms our model's ability to focus on vital cooperative cues, with 30\% providing the optimal balance.

\subsection{Qualitative Evaluation}

\begin{figure}[t] 
  \centering 
  \includegraphics[width=0.95\linewidth]{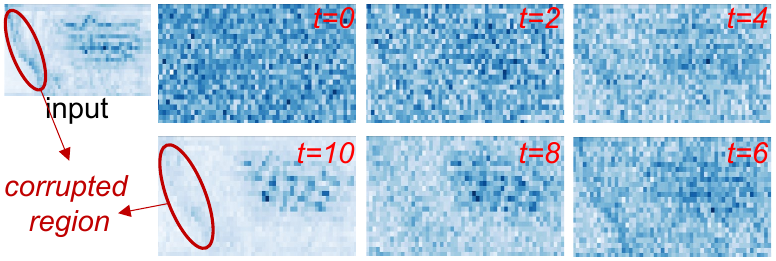}
  \caption{Visualization of the denoising process at inference time.}
  \label{fig:diff_vis} 
\end{figure}

\noindent \textbf{Diffusion Process Visualization.} 
Fig.~\ref{fig:diff_vis} illustrates the denoising trajectory of our diffusion process from $t{=}0$ (initial noise) to $t{=}10$. 
As the iteration proceeds, the corrupted regions in the input are gradually restored, showing that our model can effectively remove structured noise while preserving underlying geometric details. 

\begin{figure}[t] 
  \centering 
  \includegraphics[width=0.95\linewidth]{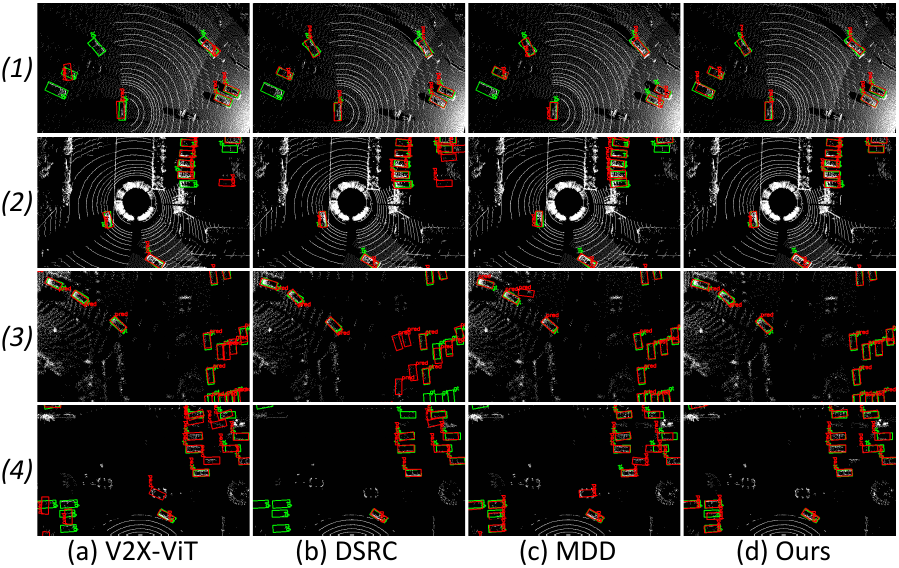}
  \caption{Visualization of detection under various corruptions, where (1) \emph{beam\_missing}, (2) \emph{fog}, (3) \emph{motion\_blur}, and (4) \emph{water}.}
  \label{fig:det_vis} 
\end{figure}

\noindent \textbf{Detection Visualization.} 
Fig.~\ref{fig:det_vis} presents qualitative comparisons under four corruption types. 
Compared with other mthods, our method produces more complete and accurate bounding boxes, especially in heavily degraded regions where baselines either miss objects or yield false positives. 
These results visually confirm the effectiveness of our diffusion-based framework in suppressing noise and recovering reliable detections under diverse corruptions.

\section{Conclusion}
We presented a diffusion-guided cooperative perception framework that enhances robustness under diverse corruptions. 
By integrating a quality-aware teacher-student design and a dual-branch gated fusion mechanism, our method effectively suppresses noise accumulation and preserves reliable cross-agent information. 
Experiments on OPV2Vn and DAIR-V2Xn demonstrate consistent improvements over state-of-the-art baselines in both accuracy and stability.
Future work will focus on mitigating the risk of diffusion models generating geometries or objects that may not be perfectly aligned with the ground truth.

\noindent \textbf{Acknowledgement.} The work is supported in part by the S\&T Program of Hebei Province (Beijing-Tianjin-Hebei Collaborative Innovation Special Program) under Grant 25240701D.


{
    \small
    \bibliographystyle{ieeenat_fullname}
    \bibliography{main}
}

\end{document}